\documentclass[10pt,twocolumn]{article}
\usepackage{booktabs}
\usepackage{float}
\restylefloat{table}
\usepackage{tabu}
\usepackage{graphics}
\usepackage{color}

\newcommand\blfootnote[1]{%
  \begingroup
  \renewcommand\thefootnote{}\footnote{#1}%
  \addtocounter{footnote}{-1}%
  \endgroup
}

\usepackage[utf8]{inputenc}
\usepackage{url}
\usepackage{graphicx}
\usepackage{authblk}

\usepackage{algorithm}
\usepackage{algorithmic}

\usepackage{hyperref}

\title{A Deep Learning Approach Towards Prediction of Faults in Wind Turbines}
\author{Joyjit Chatterjee\thanks{Corresponding Author: j.chatterjee-2018@hull.ac.uk}}
\author{Nina Dethlefs}
\affil{University of Hull, United Kingdom}

\date{\vspace{-5ex}}

\begin{document}
\raggedbottom

\maketitle

\section{Introduction}
\blfootnote{Northern Lights Deep Learning Workshop, Tromso, Norway, January 2019.}
With the rising costs of conventional sources of energy, the world is moving towards sustainable energy sources including wind energy. Wind turbines consist of several electrical and mechanical components and experience an enormous amount of irregular loads, making their operational behaviour at times inconsistent. Operations and Maintenance (O\&M) is a key factor in monitoring such inconsistent behaviour of the turbines in order to predict and prevent any incipient faults which may occur in the near future. 

Machine learning has been applied to the domain of wind energy over the last decade for analysing, diagnosing and predicting wind turbine faults. In particular, we follow the idea of modelling a turbine's performance as a power curve where any power outputs that fall off the curve can be seen as performance errors. Existing work using this idea \cite{paper1} has used data from a turbine's Supervisory Control \& Acquisition (SCADA) system to filter and analyse fault \& alarm data using regression techniques. In \cite{paper2}, the authors showed that 
traditional supervised learning techniques, e.g. support vector regression, are able to make reliable predictions from a power curve.

In this study, we investigate the applicability of deep learning techniques to turbine fault prediction from power curves. Deep learning \cite{Goodfellow-et-al-2016} is relatively new in its application to wind energy. We use the open access NREL Western Wind Dataset\footnote{\tiny \url{https://www.nrel.gov/grid/western-wind-data.html}} for the year 2012 to first predict the wind power output from turbines and secondly identify operational faults based on data points that fall off the optimal performance curve. In contrast to previous work, we explore how deep learning can be applied to fault prediction from open access meteorological data only.







\begin{figure} 
    \centering
    \includegraphics[scale = 0.2]{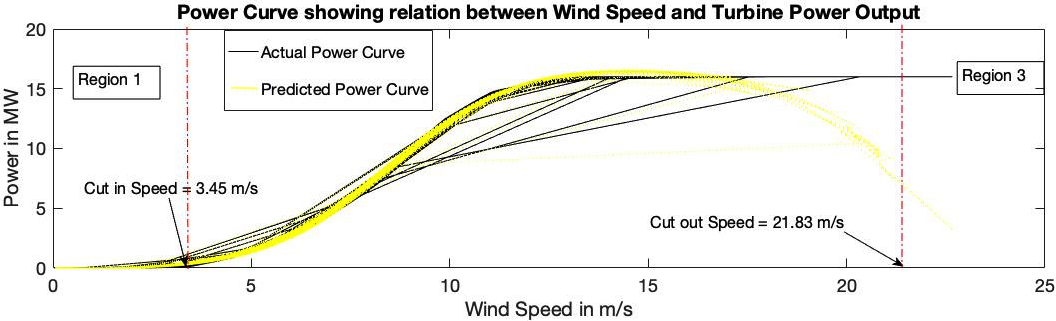}
    \caption{Power Curve for the Wind Turbine where any instance in Region 1 or Region 3 signifies a fault. Actual power curve is modelled based on the labelled data (black) while the predicted power (yellow) is modelled using a neural network}
    \label{fig:powercurve_neuralnet}
\end{figure}

\section{Modelling of Power Curve}
A power curve is an integral part of wind turbines' O\&M, showing the relation between the wind speed and the turbine's power output. A turbine is said to be operating normally when the wind speed is above the cut-in speed---which is the minimum wind speed at which the turbine generates power, or when the wind speed is below the cut-out speed---which is the maximum wind speed, beyond which turbine should stop operating to prevent failures. The ideal turbine operates at the rated speed---which is the speed at which turbine generates its rated power. Whenever a turbine enters regions 1 and 3 of the power curve (i.e. below cut-in wind speed or above cut-out), it can be interpreted as an impending fault in the turbine. See Figure \ref{fig:powercurve_neuralnet} for an illustration.

In the absence of openly available SCADA data on turbine faults, we use the 29,736 data samples in the NREL database to model faults based on the power curve and meteorological data only. Features include timestamps (month, day, hour, min), wind speed (m/s), air temperature (Kelvin), air pressure (Pascals), wind direction (Degrees) and density at hub height (Kg/$ m^3 $). We use a 70\%-30\% split of the data set of training and testing and find that a Medium Gaussian Kernel based Support Vector Regression (SVR) model achieves a minimal square error (MSE) of 0.098063 using 5-fold cross-validation. This is a much better MSE obtained than \cite{paper2}, in which SVR with a radial basis function Kernel yields an MSE of 0.6696.



Figure \ref{fig:powercurve_neuralnet} shows the resulting power curve including predicted faults. It can clearly be enunciated from the graph that the actual and predicted power is quite close, and the actual power curve resembles the predicted power curve in most of the context. The deviations from the curve, which are demonstrated by the curve showing extreme asymptotic spikes pertains to the turbine moving over to above cut-out speed (Region 3) or below cut-in speed (Region 1).

\section{Prediction of Faults}
We compare various neural network (NN) models for identifying and predicting faults in the NREL dataset. We treat fault situations as those where the turbine operates in Region 1 or 3 of the power curve. Accordingly, the entire dataset is annotated with binary labels---1 for a fault and 0 for normal operation. The dataset is split into training (70\%), validation (15\%) and test (15\%) sets and we use MSE as an optimisation criterion and early stopping. Table \ref{tab:results} shows a comparison of neural networks, alongside performance parameters. We compared a feedforward NN, recurrent NN, convolutional NN, sparse autoencoder and dynamic time series Non Linear Autoregressive (NAR) NN. Results show that the recurrent neural net performs best (by 10\%) but needs longest to train (by 9\%) compared to sparse autoencoder (2\textsuperscript{nd} best).

\begin{table}
\centering
\resizebox{\columnwidth}{!}{%
\begin{tabular}{r|lll}
\toprule
Neural Network Type                                 & Epochs & Time & MSE \\ \midrule
Feedforward Network                                &114        &15.00 sec       &0.048979    \\
Recurrent Neural Network (RNN)              &138  &22.83 sec      &0.026299     \\
Convolutional Neural Network (CNN)        &129        &18.00 sec           &0.047950     \\
Sparse Autoencoder                                  &135        &20.53 sec      &0.029314     \\
Dynamic Time Series Non Linear Autoregressive (NAR) &131        &19.18 sec      &0.031452     \\ \bottomrule
\end{tabular}%
\caption {\label{tab:results} Various Neural Network Models used and their performance evaluation}
}
\end{table}  


\section{Conclusion and Future Work}
As is evident from existing literature most work on O\&M of wind turbines is based on traditional machine learning algorithms but has not yet explored deep learning. We present a feasibility study demonstrating that deep learning is promising for application in the wind energy domain, outperforming traditional regression models. In addition, we have explored the possibility of predicting impeding turbine faults from meteorological data --- thus circumventing the problem that machine fault data is often treated as commercially sensitive and therefore not readily openly available. In future work we seek to obtain such data and confirm our results, paving the way for predicting incipient faults in the electrical and mechanical components of wind turbines.





\paragraph{Acknowledgement} We are thankful to the University of Hull's High Performance Computing facility for providing us access to MATLAB on Viper. 

\bibliographystyle{abbrv}
\bibliography{references}

\begin{thebibliography}{1}

\bibitem{paper1}
K.~L. et~al.
\newblock Diagnosing wind turbine faults using machine learning techniques
  applied to operational data.
\newblock In {\em International Conference on Prognostics and Health Management
  (ICPHM)}, Ottawa, ON, Canada, June 2016. IEEE.

\bibitem{Goodfellow-et-al-2016}
I.~Goodfellow, Y.~Bengio, and A.~Courville.
\newblock {\em Deep Learning}.
\newblock MIT Press, 2016.

\bibitem{paper2}
P.~Zhao, J.~Xia, Y.~Dai, and J.~He.
\newblock Wind speed prediction using support vector regression.
\newblock In {\em IEEE Conference on Industrial Electronics and Applications},
  Taichung, Taiwan, June 2010. IEEE.

\end{thebibliography}

\end{document}